\pdfoutput=1
\documentclass[11pt]{article}

\usepackage[final]{acl}
\usepackage{listings}
\usepackage{xcolor}
\usepackage{fvextra}
\usepackage{caption}
\usepackage{placeins}

\usepackage{times}
\usepackage{latexsym}
\usepackage{algorithm}
\usepackage{algpseudocode}
\usepackage[T1]{fontenc}

\usepackage[utf8]{inputenc}

\usepackage{microtype}

\usepackage{inconsolata}

\usepackage{graphicx}
\usepackage{tikz}
\usetikzlibrary{arrows.meta,calc,positioning}
\title{From Tables to Quantified Statements: Evaluating LLM Inference Generation through Executable Verification}

\author{
  \textbf{Mai Mohamed Eida},
  \textbf{Gunjan Anand},
  \textbf{Ayush Singh},
  \textbf{Aleksandre Maskharashvili}
\\
  University of Illinois Urbana Champaign \\
  \texttt{\{maimm2, gunjana2, ayushs13, am172\}@illinois.edu}
}

\begin{document}
\maketitle

\begin{abstract}
LLMs can generate fluent descriptions from tables, but their outputs may remain logically unsupported by the structured data. We introduce \textsc{Stat-to-Text}\footnote{\textsc{Stat-to-Text} Full code and results are released on \href{https://github.com/maimm2/Stat-To-Text}{GitHub}.}, a controlled task in which LLMs generate quantified natural language inferences from statistical tables using quantified constructions such as \textit{all}, \textit{some}, \textit{no}, and \textit{most}. To evaluate these inferences, we use an LLM generated Python checker code which when executed verifies the corresponding truth conditions against the table. We compare four open-weight LLMs across model families and scales, evaluating faithfulness, logical accuracy, table coverage, and diversity. Our results show that model scale and family matter, with the largest model (GPT-OSS-120B) consistently producing the most faithful inferences without sacrificing greater table coverage and quantifier diversity, as opposed to smaller models. These findings are supported by human annotation, which shows that the automated checker closely aligns with human judgments.
\end{abstract}

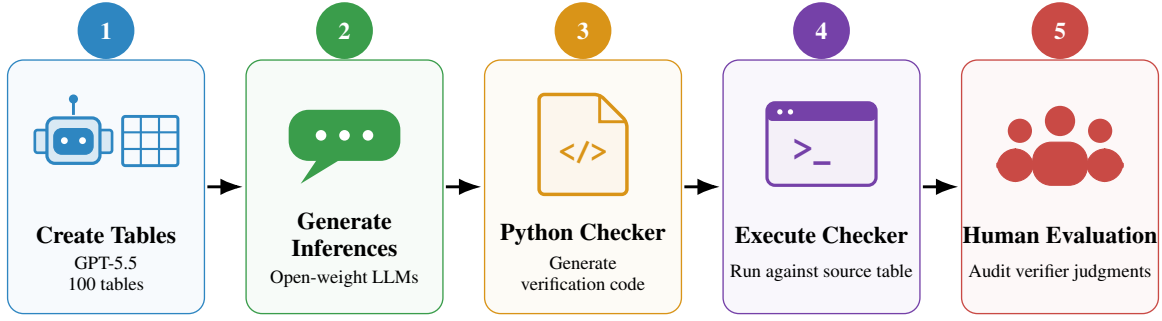
\begin{figure*}[t]
\centering

\definecolor{stageblue}{HTML}{2E86C1}
\definecolor{stagegreen}{HTML}{3A9D4B}
\definecolor{stageorange}{HTML}{D99418}
\definecolor{stagepurple}{HTML}{7541A8}
\definecolor{stagered}{HTML}{C94A45}

\begin{tikzpicture}[
    stage/.style={
        rounded corners=6pt,
        minimum width=2.6cm,
        minimum height=3.35cm,
        line width=0.5pt
    },
    stepnumber/.style={
        circle,
        minimum size=7.5mm,
        text=white,
        font=\bfseries\small,
        inner sep=0pt
    },
    stagelabel/.style={
        font=\bfseries\small,
        align=center
    },
    stagesubtitle/.style={
        font=\scriptsize,
        align=center
    },
    pipelinearrow/.style={
        -{Latex[length=3mm,width=2.2mm]},
        line width=1.1pt,
        black
    }
]


\node[
    stage,
    draw=stageblue,
    fill=stageblue!4
] (stage1) at (0,0) {};

\node[
    stage,
    draw=stagegreen,
    fill=stagegreen!4
] (stage2) at (3.15,0) {};

\node[
    stage,
    draw=stageorange,
    fill=stageorange!4
] (stage3) at (6.30,0) {};

\node[
    stage,
    draw=stagepurple,
    fill=stagepurple!4
] (stage4) at (9.45,0) {};

\node[
    stage,
    draw=stagered,
    fill=stagered!4
] (stage5) at (12.60,0) {};


\node[
    stepnumber,
    fill=stageblue
] at ($(stage1.north)+(0,0.38)$) {1};

\node[
    stepnumber,
    fill=stagegreen
] at ($(stage2.north)+(0,0.38)$) {2};

\node[
    stepnumber,
    fill=stageorange
] at ($(stage3.north)+(0,0.38)$) {3};

\node[
    stepnumber,
    fill=stagepurple
] at ($(stage4.north)+(0,0.38)$) {4};

\node[
    stepnumber,
    fill=stagered
] at ($(stage5.north)+(0,0.38)$) {5};


\draw[pipelinearrow]
    ($(stage1.east)+(0.03,0)$) --
    ($(stage2.west)+(-0.03,0)$);

\draw[pipelinearrow]
    ($(stage2.east)+(0.03,0)$) --
    ($(stage3.west)+(-0.03,0)$);

\draw[pipelinearrow]
    ($(stage3.east)+(0.03,0)$) --
    ($(stage4.west)+(-0.03,0)$);

\draw[pipelinearrow]
    ($(stage4.east)+(0.03,0)$) --
    ($(stage5.west)+(-0.03,0)$);


\begin{scope}[shift={($(stage1.center)+(0,0.55)$)}]

    \draw[
        stageblue,
        line width=1.1pt
    ] (-0.43,0.40) -- (-0.43,0.57);

    \fill[stageblue]
        (-0.43,0.61) circle (0.06);

    \draw[
        stageblue,
        fill=stageblue!18,
        line width=1pt,
        rounded corners=1pt
    ] (-0.94,-0.10) rectangle (-0.77,0.20);

    \draw[
        stageblue,
        fill=stageblue!18,
        line width=1pt,
        rounded corners=1pt
    ] (-0.09,-0.10) rectangle (0.08,0.20);

    \draw[
        stageblue,
        fill=stageblue!18,
        line width=1.2pt,
        rounded corners=3pt
    ] (-0.78,-0.24) rectangle (-0.08,0.36);

    \draw[
        stageblue,
        fill=stageblue,
        rounded corners=3pt
    ] (-0.69,-0.12) rectangle (-0.17,0.22);

    \fill[white] (-0.55,0.06) circle (0.045);
    \fill[white] (-0.31,0.06) circle (0.045);

    \draw[
        stageblue,
        line width=1.1pt,
        rounded corners=1pt
    ] (0.22,-0.27) rectangle (0.95,0.38);

    \draw[stageblue, line width=0.8pt]
        (0.46,-0.27) -- (0.46,0.38);

    \draw[stageblue, line width=0.8pt]
        (0.70,-0.27) -- (0.70,0.38);

    \draw[stageblue, line width=0.8pt]
        (0.22,-0.05) -- (0.95,-0.05);

    \draw[stageblue, line width=0.8pt]
        (0.22,0.17) -- (0.95,0.17);

\end{scope}

\node[
    stagelabel
] at ($(stage1.center)+(0,-0.63)$)
    {Create Tables};

\node[
    stagesubtitle
] at ($(stage1.center)+(0,-1.14)$)
    {GPT-5.5\\100 tables};


\begin{scope}[shift={($(stage2.center)+(0,0.58)$)}]

    \draw[
        stagegreen,
        fill=stagegreen,
        line width=1pt,
        rounded corners=7pt
    ]
        (-0.72,-0.22)
        rectangle
        (0.72,0.42);

    \fill[
        stagegreen
    ]
        (-0.35,-0.20) --
        (-0.58,-0.54) --
        (-0.04,-0.21) --
        cycle;

    \fill[white] (-0.31,0.10) circle (0.07);
    \fill[white] (0,0.10) circle (0.07);
    \fill[white] (0.31,0.10) circle (0.07);

\end{scope}

\node[
    stagelabel
] at ($(stage2.center)+(0,-0.63)$)
    {Generate\\Inferences};

\node[
    stagesubtitle
] at ($(stage2.center)+(0,-1.22)$)
    {Open-weight LLMs};


\begin{scope}[shift={($(stage3.center)+(0,0.56)$)}]

    \draw[
        stageorange,
        fill=stageorange!8,
        line width=1.3pt,
        rounded corners=2pt
    ]
        (-0.58,-0.66) --
        (-0.58,0.66) --
        (0.22,0.66) --
        (0.58,0.30) --
        (0.58,-0.66) --
        cycle;

    \draw[
        stageorange,
        line width=1.1pt
    ]
        (0.22,0.66) --
        (0.22,0.30) --
        (0.58,0.30);

    \node[
        text=stageorange,
        font=\ttfamily\bfseries\large
    ] at (0,-0.10)
        {\textless/\textgreater};

\end{scope}

\node[
    stagelabel
] at ($(stage3.center)+(0,-0.63)$)
    {Python Checker};

\node[
    stagesubtitle
] at ($(stage3.center)+(0,-1.14)$)
    {Generate \\ verification code};


\begin{scope}[shift={($(stage4.center)+(0,0.56)$)}]

    \draw[
        stagepurple,
        fill=stagepurple!8,
        line width=1.2pt,
        rounded corners=3pt
    ]
        (-0.72,-0.54)
        rectangle
        (0.72,0.55);

    \fill[
        stagepurple,
        rounded corners=2pt
    ]
        (-0.72,0.31)
        rectangle
        (0.72,0.55);

    \fill[white] (-0.55,0.43) circle (0.035);
    \fill[white] (-0.41,0.43) circle (0.035);

    \node[
        text=stagepurple,
        font=\ttfamily\bfseries\Large
    ] at (-0.12,-0.10)
        {\textgreater\_};

\end{scope}

\node[
    stagelabel
] at ($(stage4.center)+(0,-0.63)$)
    {Execute Checker};

\node[
    stagesubtitle
] at ($(stage4.center)+(0,-1.14)$)
    {Run against source table};


\begin{scope}[shift={($(stage5.center)+(0,0.54)$)}]

    \fill[stagered] (0,0.33) circle (0.20);
    \fill[stagered] (-0.53,0.20) circle (0.16);
    \fill[stagered] (0.53,0.20) circle (0.16);

    \draw[
        stagered,
        fill=stagered,
        rounded corners=7pt
    ]
        (-0.36,-0.52)
        rectangle
        (0.36,0.05);

    \draw[
        stagered,
        fill=stagered,
        rounded corners=7pt
    ]
        (-0.83,-0.47)
        rectangle
        (-0.35,-0.04);

    \draw[
        stagered,
        fill=stagered,
        rounded corners=7pt
    ]
        (0.35,-0.47)
        rectangle
        (0.83,-0.04);

\end{scope}

\node[
    stagelabel
] at ($(stage5.center)+(0,-0.63)$)
    {Human Evaluation};

\node[
    stagesubtitle
] at ($(stage5.center)+(0,-1.14)$)
    {Audit verifier judgments};

\end{tikzpicture}

\caption{\textsc{Stat-to-Text}: Quantified inference generation and
evaluation pipeline. An LLM generates both quantified inferences and Python checker scripts based on statistical tables and inferences respectively. Python checkers scripts are executed on the source tables, and a sampled subset is evaluated by human annotators to validate the LLM inference verification output.}
\label{fig:pipeline}
\end{figure*}
\section{Introduction}

Data-to-text generation aims to render structured inputs, such as tables \cite{wang:2021:SAR}, graphs \cite{song:2020:SIP}, records, and meaning representations, as accurate and faithful natural language~\citep{reiter2000building,lin:2024:ASO}. Traditional Natural Language Generation (NLG) pipelines decompose generation into content selection, sentence planning, and surface realization, whereas recent neural language models integrate these stages through end-to-end \cite{yin:2022:seq2seq}, fine-tuning~\cite{gong:2019:ETM}, or prompt-based generation.

In this work, we focus on table-to-text generation, where the input is a table containing numerical and categorical variables.
Large language models (LLMs) now achieve strong performance on
structured data-to-text tasks without task specific fine-tuning or
large collections of human written references~\cite{zhao:2023:ITTT}.
In-context learning and chain-of-thought prompting can further
support structured inputs through reasoning~\citep{wei:2022:COT,zhao:2023:ITTT}.
Nevertheless, fluent outputs may be unsupported by the source table,
overlook relevant rows or variables, overgeneralize from a small
subsets of records, misrepresent numerical relations, or repeat
near-identical claims. Faithfulness, hallucinations, factual accuracy, coverage, and diversity therefore remain central challenges in LLM based table-to-text generation ~\citep{calo-etal-2026-logic, kasner-etal-2021-text}.

Although prior table-to-text research has incorporated numerical and logical operations ~\citep{liu:2021:TFI,zhao:2023:ITTT,trigg:2025:LTTT} such as aggregation, comparison, superlatives, and majority reasoning ~\citep{suadaa:2021:numericnlg,moosavi:2021:scigen,chen:2020:LNL,chen:2020:logic2text,trigg:2025:LTTT}, generalized quantifier selection has received comparatively limited attention as a primary generation and evaluation objective. A generalized quantifier, such as \textit{all}, \textit{some}, \textit{no}, \textit{most}, \textit{at least half}, \textit{more than two-thirds}, \textit{exactly three}, and \textit{all but one}~\citep{Barwise1981-BARGQA}, expresses a relation between two sets: the \textit{restrictor} ($R$), which defines the domain of quantification, and the \textit{nuclear scope} (henceforth \textit{scope} -- $S$), which specifies the property attributed to that domain. Generalized quantifiers are particularly well suited to statistical table-to-text generation as they can represent patterns spread across the table as brief, understandable natural language statements. This enables users to grasp overall statistical observations without examining each table row individually. At the same time, they impose precise semantic constraints that can be formally verified: for example, in 
\textit{Most people over 70 are self-employed or retired} the standard interpretation of \textit{most} requires more than half of $R$ (\textit{people over 70}) be members of $S$ (\textit{self-employed or retired}). Generating such statements therefore calls for filtering, counting, comparison, aggregation, and quantifier selection. To the best of our knowledge, prior work has not systematically examined whether LLM generated quantified inferences are supported by a fixed statistical table or whether models distinguish the truth conditions associated with generalized quantifiers.

In this paper we focus on a form of table-to-text generation, which we call \textsc{Stat-to-Text}: generating faithful quantified natural language statements inferred from a \textit{statistical table}, i.e., table of numerical and categorical variables from which meaningful statistics can be drawn. The target output is not a general summary or a verbalization of selected cells, but a set of quantified statements over table defined row sets. For example, given a table containing ``age'' and ``employment'' status, a model may generate \textit{Most people over 70 are self-employed or retired}. After generating inferences, our pipeline, illustrated in Figure~\ref{fig:pipeline}, uses an LLM generated Python checker to verify the faithfulness of the inferred statements. Then we evaluate the quality of the results quantitatively by checking faithfulness, coverage, and diversity of the generated statements, followed by a qualitative evaluation through human annotation. In this work, we address the following research questions:

\textbf{RQ1.} \textit{How reliably does our \textsc{Stat-to-Text} pipeline generate and verify quantified natural language inferences from statistical tables?}

\textbf{RQ2.} \textit{How do model family, model scale, and reasoning paradigm (direct prediction vs. reasoning models) affect the logical faithfulness of quantified \textsc{Stat-to-Text} generation?}

\textbf{RQ3.} \textit{How do LLMs differ in the coverage and diversity of the quantified inferences they generate?}

Our results show that the \textsc{Stat-to-Text} pipeline reliably generates and verifies faithful quantified inferences from statistical tables, with the limitation of not yet achieving 100\% throughput, as some tables are dropped during inference generation or checker execution. We further find that both model family and model scale influence logical faithfulness, with GPT-OSS-120B achieving the highest faithfulness without sacrificing overall table and quantifier coverage, and with reasonable variable and quantifier diversity. This is strengthened by GPT-OSS-120B maintaining a near perfect agreement with human annotators (97.5\%), the highest of all models.
 
\section{Related Work}
 
\subsection{Table-to-Text Generation and Table Reasoning}

Table-to-text generation has evolved from modular NLG pipelines to neural and LLM-based systems that generate directly from tabular inputs~\citep{lin:2024:ASO, liu:2022:PLOG}. Early neural benchmarks emphasized fluent generation and content selection ~\cite{castro:2019:NDTT}, while later work increasingly focused on factual grounding and reasoning over multiple table records, such as RotoWire~\cite{wiseman:2017:challenges} and ToTTo \cite{parikh:2020:totto}. Both benchmarks focus on generating natural language descriptions from table content rather than generating analytical or quantified inferences.

Recent work extends table-to-text generation from cell verbalization toward numerical inference \citep{onderkova:2025:FRESHTAB}. Both NumericNLG~\citep{suadaa:2021:numericnlg} and SciGen~\citep{moosavi:2021:scigen} focus on descriptions of numerical reasoning over values in tables from scientific papers, where such descriptions require arithmetic and comparative reasoning. In contrast, approaches such as \citet{perez:2025:ALBA} generate textual insights from multi-table databases through an intermediate SQL-based reasoning step, rather than by directly prompting an LLM over the table. 

Logical table-to-text generation complements this numerical perspective. LogicNLG ~\citep{chen:2020:LNL} and  Logic2Text ~\citep{chen:2020:logic2text} formulate generation as producing natural language statements that are logically entailed by open-domain semistructured tables. They pair table grounded descriptions with logical forms and cover operations such as comparison, aggregation, superlatives, majority, and uniqueness. These tasks are closely related to ours, but they treat such relations as part of a broader inventory of logical operations. We instead isolate generalized quantifier selection and evaluate whether the generated quantifier correctly expresses the relation over the table.

LLMs have further broadened table-to-text generation beyond fixed benchmark settings. \citet{zhao:2023:ITTT} evaluate LLMs as generators, evaluators, and feedback providers for table insight and query-based generation tasks, while \citet{trigg:2025:LTTT} survey logical table-to-text generation. This work shows that prompted LLMs can perform flexible table interpretation without task specific training, and identify reasoning, numeracy, faithfulness, and verification as central challenges. While these studies motivate the use of LLMs for table based inferences, it remains open how reliably models generate quantified claims whose truth depends on relations distributed across multiple rows.

\subsection{Generalized Quantifiers}
 
Generalized quantifiers are analyzed in formal semantics as relations between sets~\citep{Barwise1981-BARGQA}. For our running example, \textit{Most people over 70 are self-employed or retired}, the standard interpretation of \textit{most} requires more than half of $R$ belong to $S$:

$
\qquad \mathrm{most}(R,S) \iff |R \cap S| > \frac{1}{2}|R|
$
\\
This interpretation maps naturally onto tabular data. Rows define entities; headers serve as attributes used to construct predicates on numerical and categorical variables, giving rise to restrictors and scopes. Proportional quantifiers such as \textit{most} and \textit{at least half} depend on the ratio of $|R \cap S|$ to $|R|$; universal quantifiers such as \textit{all} require $R \subseteq S$; existential quantifiers such as \textit{some} require $R \cap S \neq \emptyset$; negative quantifiers such as \textit{no} require $R \cap S = \emptyset$; and cardinal quantifiers such as \textit{exactly three} depend on the size of $R \cap S$. Generating a faithful quantified statement therefore means identifying the restrictor and scope predicates, computing the corresponding set relation, and selecting a linguistic expression with matching truth conditions.
 
When it comes to quantified statements in NLG, \citet{chen:2019:generating} study quantified descriptions of abstract visual scenes and evaluate whether generated descriptions are correct, complete, and human-like. We transfer this problem to tabular data, where the restrictor and scope must be constructed from numerical and categorical table attributes and the selected quantifier must be licensed by the corresponding row counts. 
 
\subsection{LLM Python Verification}
Faithfulness remains a central limitation of table-to-text generation. This gap has been addressed through entity centric perspectives~\cite{liu:2021:TFI}, token level detection of incorrect values and statements~\cite{kasner-etal-2021-text}, and logical consequence of unsupported outputs~\cite{calo-etal-2026-logic}.

However, another method to address faithfulness is through Program-aided language models (PAL)~\citep{gao:2023:PAL}, where it delegates
symbolic and numerical operations to executable code. In this paper, we use PAL to rely on execution-based reasoning, and we adapt this idea to evaluation, translating each generated inference into a Python verification function that is executed against the source table to address faithfulness. The reliability of other automatic table-to-text evaluation is an open problem: automatic metrics do not always agree with LLM-based or human judgments of table generated insights~\citep{onderkova:2025:FRESHTAB}. This motivates evaluating the semantic content using PAL of individual statements rather than relying on exclusively learned metrics which are not suitable for \textsc{Stat-to-Text} evaluation. Our work therefore connects numerical table-to-text generation, logical table reasoning, quantified expression generation, and factuality evaluation. 

\section{Data Setup}
Since our experiments require previously unseen tables with controlled row and header counts, we prompt GPT-5.5 to generate a synthetic dataset of tabular data. We select a different model for table generation than those used in the next stages of our pipeline to ensure consistency across all synthetic tables while avoiding potential bias toward any of the models evaluated in our experiments.

We generate 100 tables, each containing 5 - 10 variables and 15 rows, using the prompt illustrated in Appendix~\ref{sec:fullprompts}. The topics of these tables range across the following themes: healthcare, retail/business, transportation/logistics, household finance, sports, agriculture, and HR/employee management, as illustrated in Table~\ref{tab:table_synthetic_example_short}. A full sample table is shown in Table~\ref{tab:table_synthetic_example} 
in Appendix~\ref{sec:synthetic_tables}. 
\begin{table*}[t]
\centering
\small
\setlength{\tabcolsep}{6pt}
\renewcommand{\arraystretch}{1.15}
\resizebox{\textwidth}{!}{%
\begin{tabular}{|l|l|c|c|c|c|c|c|}
\hline
\textbf{Patient ID} & \textbf{Diagnosis} & \textbf{Age} & \textbf{Systolic BP} & \textbf{Diastolic BP} & \textbf{Chol.} & \textbf{BMI} & \textbf{Smoker} \\
\hline
PT001001 & asthma       & 46 & 129 & 72 & 192 & 33.3 & no \\
\hline
PT001002 & diabetes     & 62 & 143 & 82 & 223 & 22.5 & yes \\
\hline
PT001003 & diabetes     & 68 & 147 & 87 & 198 & 31.0 & no \\
\hline
PT001004 & arthritis    & 46 & 135 & 77 & 182 & 27.6 & yes \\
\hline
PT001005 & hypertension & 76 & 119 & 74 & 245 & 22.7 & no \\
\hline
\end{tabular}}
\caption{Example of a synthetic statistical table generated by GPT-5.5.}
\label{tab:table_synthetic_example_short}
\end{table*}

Fixing rows and number of variables allows us to control for table complexity, ensuring the evaluation results can be attributed to the models rather than variations in the source table size. 

\section{\textsc{Stat-to-Text} Pipeline Setup}
\paragraph{Step 1 -- LLM Inferences:} After generating our statistical tables, we prompt all selected LLMs to generate quantified inferences for each statistical table. We limit the inferences to 20 inferences, and specify a diverse set of universal, existential, majority, negation, and conditional  statements in the few-shot prompting. The same prompt is used across both direct prediction models and reasoning models (Figure~\ref{fig:statement_prompt} in Appendix~\ref{sec:fullprompts}).  Therefore, models generate inferences similar to the following (more examples in Appendix~\ref{sec:inference&CheckerOutput}):
 \begin{quote}
 \small
 \begin{enumerate}
     \item All individuals in the table have a systolic blood pressure between 112 and 156 mmHg.
      \item For all individuals with a BMI greater than 30, their age is greater than 40 years.
 \end{enumerate}
\end{quote}

 Specifying a fixed number of generated inferences is crucial at this stage to preserve inference quality, otherwise, some models generate 100+ inferences until the token limit is reached, despite explicit instruction to only generate 20 meaningful inferences. However, some reasoning models skip inference generation after becoming stuck in the reasoning process and exhausting the available tokens, dropping tables in the process. After generating 20 inferences for all the tables in the dataset, we move to step 2 in the pipeline: an LLM generated Python Checker.

\paragraph{Step 2 -- LLM Generated Python Checker:} We prompt the model to generate a Python script that automatically verifies the factual correctness of each generated inference with respect to its corresponding table. In the prompt, we provide the table header (variables) and the list of inferences generated in step 1 for each table. We omit the table cell values to prevent contaminating the script generation step from relying on specific values rather than producing a generalizable verification script.

The script contains one verification function per inference, where each function evaluates whether the statement is true or false by querying the table using the pandas library. In addition to the binary verdict, the script outputs a brief justification for its decision, as illustrated in Listing~\ref{lst:verification}. This automated verification allows us to assess the factual consistency of the generated inferences, while ensuring that every inference is evaluated using the same procedure. 

\begin{lstlisting}[
caption={Excerpt of an automatically generated Python verification function.},
captionpos=b,
label={lst:verification}
]
def stmt_1(df: pd.DataFrame):
    """All individuals have systolic blood pressure
    between 112 and 156 mmHg."""

    local = df.copy()
    local["bp_systolic"] = pd.to_numeric(
        local["bp_systolic"], errors="coerce"
    )

    mask = local["bp_systolic"].between(112, 156)
    truth = bool(mask.all())

    if truth:
        return True, "No violations found."
    else:
        return False, (
            f"Found {len(local[~mask])} violating rows."
        )
\end{lstlisting}

In this stage, we developed two prompt variants to account for differences in prompt sensitivity between direct-prediction models and reasoning models. Reasoning models generally produced valid verification scripts from concise prompts~(Figure~\ref{fig:initial_code_generation_prompt} in Appendix~\ref{sec:fullprompts}). However, direct-prediction models often required additional implementation constraints to improve consistency and reduce common code generation errors across models (Figure~\ref{fig:code_generation_prompt_LLAMA} in Appendix~\ref{sec:fullprompts}).

\section{Model Selection}
We evaluate four open-weight LLMs spanning two model families and two parameter scales, including both direct-prediction and reasoning-oriented models:  LLAMA 3.1 8B, LLAMA 3.1 70B~\cite{grattafiori:2024:llama}, GPT-OSS 20B, and GPT-OSS 120B~\cite{openai2025gptoss120bgptoss20bmodel}. These models were selected to examine the effects of both model family and model scale on inference generation and verification performance. We focus on open-weight models to enable reproducibility and to limit computation cost associated with repeatedly querying proprietary APIs throughout the evaluation pipeline. The only exception is GPT-5.5~\cite{openai2026gpt55}, which is used exclusively for synthetic statistical table generation, as justified in Section 3.

\section{Automated Evaluation}

At this stage, we have executed the pipeline and need to evaluate the quality of the generated inferences. We adapt previous table-to-text evaluation metrics~\cite{liu:2021:TFI, lin:2024:ASO} and introduce additional metrics tailored to quantified inference generation. For a set of inferences to be informative, they must first be faithful: grounded in the source table without referencing external information, and logically accurate with respect to the table. They must also exhibit high coverage by representing both the table headers and rows, while covering full generalized quantifier categories (e.g., universal, existential, conditional, and negation). Finally, the generated inferences should be diverse. We therefore measure both variable redundancy and quantifier redundancy to ensure that the model does not repeatedly focus on the same variables or lexical realizations of quantifiers.  Our results are illustrated in Table~\ref{tab:step-d-results-overall}.

\subsection{Faithfulness}
\textbf{Grounding} is measured through whether the inferences refer only to variables, entities, categories, and values present in the source table. \textbf{Accuracy} measures whether a generated statement is true with respect to the source table: a statement is accurate when the relation expressed by its predicates and quantifier satisfies the corresponding truth conditions over the relevant table cells. Taken together, these metrics provide a measurement for faithfulness~\cite{liu:2021:TFI}.

\textbf{Results} All models achieve near-perfect grounding (99.5--100\%), which means tables are not generating statements which references external values. This is a desirable outcome, however, taken with the accuracy results  (LLAMA-8B accuracy rate is 48.7\% despite achieving 100\% in grounding), it seems that grounding alone is insufficient to ensure logically valid (accurate) quantified inferences. Scale does matter for faithfulness, with GPT-OSS-120B achieving the best faithfulness scores.
\subsection{Coverage}
\textbf{Table Record Coverage (TRC)} quantifies the extent to which the generated inferences utilize the information contained in the rows of the input table. TRC is computed as the number of unique table rows referenced by at least one inference divided by the total number of rows in the table. A value of 1 indicates that every table row contributes to at least one generated inference, whereas lower values indicate that only a subset of the table is represented in the generated inferences.

\[
\mathrm{TRC} =
\frac{\left|\bigcup_{i=1}^{N} R_i\right|}
{|T|}
\]

where \(T\) denotes the set of all table rows, \(R_i\) is the subset of rows
referenced by the \(i\)-th inference, and \(N\) is the total number of
generated inferences.

\textbf{Results} As illustrated in Table \ref{tab:step-d-results-overall}, our results show that all models achieve high TRC, ranging from 92--100\% across the generated inferences. GPT-OSS-120B and LLAMA-70B achieve the highest coverage, reaching 100\% and 99\%, respectively. Model scale impacts row coverage, though the difference in impact is negligible.

\textbf{Variable Coverage (VC)} measures the extent to which the generated inferences utilize the variables present in the input table. A variable is captured if it is referenced by name or by a column value in at least one inference. If the inference mentions the column value ``diabetes'' it would consider the variable ``diagnosis'' as being captured. 
\[
\mathrm{VC} =
\frac{\left|\bigcup_{i=1}^{N} V_i\right|}
{|V|}
\]

where \(V\) is the set of all variables (columns) in the input table,
\(V_i\) is the subset of variables referenced by the \(i\)-th inference,
and \(N\) is the total number of generated inferences. VC, taken together with TRC, illustrates the entire coverage of the table rows and columns. Higher values indicate that the generated inferences make use of a broader range of the available information. 

\textbf{Results} VC scores vary between 87--95\%~(Table~\ref{tab:step-d-results-overall}), showing lower coverage of columns than for TRC scores for rows. However, VC closely mirrors the trends observed for TRC in relation to scale, indicating that larger models maintaining broader row coverage also preserve more comprehensive coverage of the underlying table variables with LLAMA-70B (94.8\%) and GPT-OSS-120B (95.6\%) achieving the highest VC coverage.

\begin{table*}[t]
\centering
\includegraphics[width=1\linewidth]{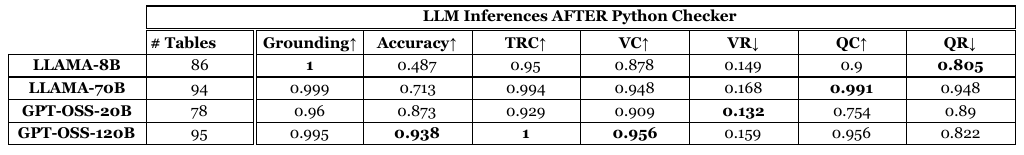}
\caption{Performance of open-weight LLMs on \textsc{Stat-to-Text} generated inferences' faithfulness (grounding and accuracy), coverage (TRC, VC, QC), and diversity (VR, QR) \textit{after} Python-based verification. Metrics annotated with $\uparrow$ indicate that higher values correspond to better performance, whereas metrics annotated with $\downarrow$ indicate that lower values are preferred. \textsc{Stat-to-Text} does not achieve 100\% throughput, as some tables are dropped during Steps 1 or 2 of the pipeline, resulting in the number of remaining tables shown in the Table column.} 
\label{tab:step-d-results-overall}
\end{table*}
\begin{figure*}[t]
        \centering
        \includegraphics[width=0.75\textwidth]
        {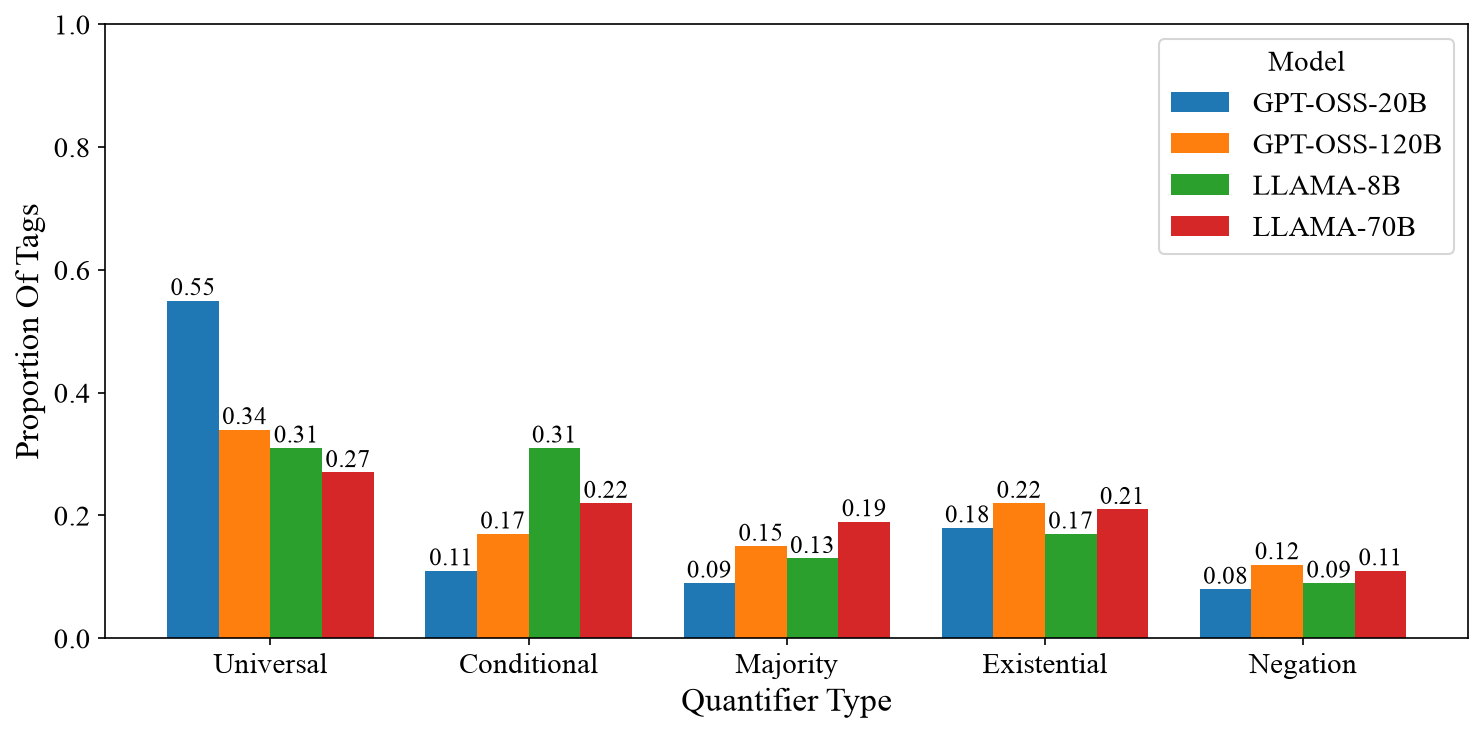}
                \caption{Distribution of quantifier categories generated by each model across the synthetic statistical tables. Bars represent the proportion of generated inferences assigned to each quantifier category (Universal, Conditional, Majority, Existential, and Negation), aggregated over all generated outputs. Values above each bar indicate the corresponding proportion.}
        \label{fig:quantifier_distribution}
\end{figure*}
\textbf{Quantifier Coverage (QC)} evaluates the use of all generalized quantifiers categories across the generated inferences. We analyze this from two perspectives. First, we evaluate Quantifier Coverage (QC), which measures whether the model uses the full categories of generalized quantifiers (Table~\ref{tab:step-d-results-overall}). Specifically, we calculate the number of distinct quantifier categories used in the 20 inferences divided by the five predefined categories (universal, existential, majority/frequency, negation, and conditional). Second, we examine the distribution of quantifier categories to determine whether models favor certain quantifiers over others. To do so, we categorize each generated inference according to its primary quantifier category and report the frequency of each category.  The resulting distributions are illustrated in Figure~\ref{fig:quantifier_distribution}.

\textbf{Results} QC, (Table~\ref{tab:step-d-results-overall}), is high across all models but is consistently highest for the LLAMA family in both coverage and distribution. This shows that the full set of generalized quantifier categories prompted is consistently represented across the generated inferences, despite the prompt providing only examples rather than explicitly requiring each quantifier category. However, higher QC does not correspond to greater faithfulness. Although the LLAMA models consistently use a wider range of quantifier categories (LLAMA-70B, QC = 99.1\%), the GPT-OSS models produce substantially more faithful quantified inferences (Grounding $\geq$ 95\% and Accuracy $\geq$ 90\%). Further, GPT-OSS-20B achieves high accuracy despite its smaller scale; however, this appears to come with a trade-off, as its quantifier distribution is more heavily skewed toward universal quantifiers~(Figure~\ref{fig:quantifier_distribution}). This suggests that generating a broader variety of quantifier categories is largely independent of accurately applying their corresponding truth conditions.

\subsection{Diversity}
\textbf{Variable Redundancy (VR)} measures the extent to which generated inferences repeatedly reference the same
table variables. While some repetition is expected, excessive reuse indicates
that the model concentrates on a smaller subset of variables instead of
generating inferences that make use of the full table. For each table \(t\),
let \(V_t\) denote the set of unique variables (columns), and let \(f_t(v)\)
denote the number of generated inferences that reference variable \(v\). We
define variable redundancy as

\[
\mathrm{VR}_t =
\frac{
    \sum_{v \in V_t} \max\left(f_t(v)-|V_t|,\,0\right)
}{
    \sum_{v \in V_t} f_t(v)
}.
\]

The first \(|V_t|\) occurrences of each variable are not considered redundant,
while each additional occurrence contributes to the numerator. We use
\(|V_t|\) as the redundancy threshold because some variable reuse is
unavoidable. For example, a table with six variables and twenty generated
inferences cannot describe every inference using entirely distinct variables,
making repeated references inevitable. By allowing each variable to appear up
to \(|V_t|\) times before counting additional occurrences as redundant, the
metric distinguishes expected reuse from excessive concentration on a small
subset of variables.

A value of \(0\) indicates that no variable is referenced more than
\(|V_t|\) times within the generated inferences for a table. Higher values
indicate greater repetition of the same variables and a stronger concentration
of the generated inferences on a smaller subset of the available table
variables. We calculate \(\mathrm{VR}_t\) separately for each table and report
the mean across all tables for each model.

\textbf{Results} VR remains low across all models~(Table \ref{tab:step-d-results-overall}), indicating that the generated inferences are generally distributed across the available table variables rather than concentrating on only a few. GPT-OSS-20B achieves the lowest redundancy (13.2\%), while LLAMA-70B exhibits the highest redundancy (16.8\%). Unlike Variable Coverage, which measures the breadth of variables represented, VR captures how evenly the generated inferences are distributed over those variables. Together, these metrics suggest a trade-off: larger models achieve broader variable coverage while exhibiting slightly higher redundancy, whereas smaller models reference variables more evenly but cover a narrower subset of the available variables.

\textbf{Quantifier Redundancy (QR)}
measures lexical concentration within each quantifier category. We use for it normalized Herfindahl–Hirschman Index. Let $L_t$ denote the set of available lexical realizations for quantifier category $t$, such as \textit{all}, \textit{every}, \textit{each}, and \textit{any} for universal quantification. If $f_{t,\ell}$ is the frequency of lexical realization $\ell$ and $N_t=\sum_{\ell \in L_t}f_{t,\ell}$, quantifier redundancy for category $t$ is defined as

\[
\mathrm{QR}_t=
\frac{
\sum_{\ell\in L_t}
(f_{t,\ell}/N_t)^2
-
1/|L_t|
}{
1-1/|L_t|
}.
\]

The metric ranges from $0$ to $1$. A score of $0$ indicates that the available lexical realizations (``all'', ``every'') are distributed evenly within each quantifier category (universal quantifiers), reflecting high lexical variety, whereas a score of $1$ indicates that a single realization accounts for all occurrences of that quantifier category. The overall quantifier redundancy score for a table is calculated as the frequency-weighted average across quantifier categories.

\textbf{Results} As illustrated in Table \ref{tab:step-d-results-overall}, Quantifier Redundancy (QR) varies considerably across models. LLAMA-8B achieves the lowest QR (80.5\%), indicating the greatest lexical variety in expressing quantifiers but this comes at the cost of faithfulness, making the quantifier diversity useless if the inferences are incorrect. GPT-OSS-120B exhibits the second lowest redundancy, while GPT-OSS-20B and LLAMA-70B rely more heavily on a smaller set of lexical realizations. 

\section{Human Evaluation}
\begin{table*}[t]
\centering
    \includegraphics[width=1\linewidth]{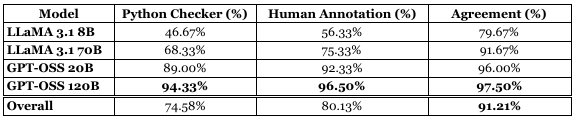}
\caption{Human evaluation random 30 tables per model. Python Checker reports the automatically computed inference accuracy, Human Annotation reports the manually annotated accuracy, and Agreement denotes the agreement between the automated and human judgments.}
\label{tab:human_eval}
\end{table*}
While the Python checker verification provides a scalable method to evaluate logical faithfulness, the scripts are generated by LLMs and therefore may contain implementation errors that misclassify inferences. To measure the reliability of \textsc{Stat-to-Text}, we manually annotate the quantified inferences generated and verified for 30 randomly sampled tables of the 100 tables from all four models. This leads to 30 tables across 4 models, with a total of 120 tables annotated. Results are illustrated in Table~\ref{tab:human_eval}. Overall, the automated verdicts agree with the human labels on 91.2\%, where we find more false negatives in the results of the pipeline than false positives. Of the 211 disagreement cases, 172 were false negatives, where the verification script labeled a human judged true inference as false, while the remaining 39 were false positives. Therefore, the automated validity scores should be interpreted as conservative lower bounds on true inference validity. Consistent with our previous evaluation, scale and family does matter, once again with the highest agreement observed with GPT-OSS-120B, and the lowest in LLAMA-3.1-8B, both in inference generation and generating Python checker verification accuracy. 
 
 During this qualitative evaluation, we observed several unexpected failures, most concentrated with the LLAMA-3.1-8B. These failures extend to verification code execution as well as generated inferences. The first involves cases where the checker is able to select the correct rows that are relevant to an inference but then assigns an incorrect ``True'' or ``False'' label while listing those rows as supporting a violation (Appendix \ref{sec:human_eval_1}). The second is related to LLMs performing a quantifier logical failure. In this case, majority/non-majority claims in which any non-satisfying row is counted as a violation, so ``most'' and ``there exists at least" is evaluated as if it was ``all'' (Appendix \ref{sec:human_eval_2}).  A third recurring failure is a verdict that contradicts its own justification: in at least 8 of the disagreements the checker's explanation reports zero violating rows, yet the label it emits is still ``false'' (Appendix \ref{sec:human_eval_3}). A fourth failure involves the checker's inability to correctly evaluate ``X or Y'' statements, instead marking the entire statement as ``False'' when one of the two conditions is not satisfied (Appendix \ref{sec:human_eval_4}). Despite this, the human evaluation supports the conclusion that execution based verification with LLM generated code is a reliable evaluation instrument when the code generating model is sufficiently capable, reaching 91.5--97.5\% agreement with human judgments for the three strongest models.

\section{Discussion}
\textbf{RQ1.} \textit{How reliably does our \textsc{Stat-to-Text} pipeline generate and verify quantified natural-language inferences from statistical tables?}

\textbf{Findings:} Our automated and human evaluation demonstrates that the \textsc{Stat-to-Text} pipeline can reliably generate and verify quantified natural-language inferences from statistical tables, provided an appropriate model is selected. Among the evaluated models, GPT-OSS-120B performs best, generating faithful quantified inferences without sacrificing coverage or diversity. Furthermore, it achieves the highest agreement with human annotations (97.5\%), demonstrating that its automatically verified inferences closely align with human judgments, concluding that our \textsc{Stat-to-Text} pipeline is reliable for quantified inference generation from statistical tables.

\textbf{RQ2.} \textit{How do model family, model scale, and reasoning paradigm (direct prediction vs. reasoning models) affect the logical faithfulness of quantified \textsc{Stat-to-Text} generation?}

\textbf{Findings:} Our results show that model family, model scale, and, in this case, reasoning paradigm have the strongest influence on logical faithfulness, with GPT-OSS-120B consistently outperforming all other evaluated models. We also observe a clear model family effect, as the GPT-OSS models produce substantially more faithful quantified inferences than the LLAMA models despite exhibiting lower quantifier diversity. The LLAMA models generate more diverse quantifier categories and achieve slightly broader variable coverage; however, these advantages do not translate into greater logical faithfulness. As quantified inferences must first be faithful before diversity becomes meaningful, the GPT-OSS models provide the strongest overall performance. As for the reasoning paradigm, further research is needed to isolate and definitively determine its contribution independently of model family and architecture. 

\textbf{RQ3.} \textit{How do LLMs differ in the coverage and diversity of the quantified inferences they generate?}
\textbf{Findings:} Our results show that models do not struggle with coverage regardless of performance on other metrics. However, diversity of quantifiers seems to come at a cost for faithfulness, especially with smaller models. With the exception of GPT-OSS-120B, models with higher faithfulness seem to exhibit lower quantifier diversity and vice versa, which will be the focus of our future research.

\section{Conclusions and Future Work}
In this paper, we present a new pipeline, \textsc{Stat-to-Text}, which reliably generates and verifies faithful quantified inferences from statistical tables. We find that while all models achieve high coverage of information from the source tables, model family and model scale have a substantial impact on the logical faithfulness of the generated inferences. Although the LLAMA models exhibit greater quantifier diversity and slightly broader variable coverage, the GPT-OSS models consistently produce more faithful quantified inferences. Finally, we find high agreement between the automated evaluation and human annotators, showing that the proposed pipeline provides a reliable framework for future quantified table-to-text generation.

Future work will focus on incorporating real world tables and improving the reliability of the Python verifier to better distinguish inference generation errors from verification errors, potentially through stronger models or supervised training on verification code. We also plan to evaluate the pipeline on larger and more complex statistical tables, where accurately verifying quantified statements becomes increasingly challenging. Finally, we aim to extend \textsc{Stat-to-Text} to coherent multi-sentence discourse inferences while preserving executable verification of individual claims and discourse relations connecting them.

\section*{Limitations}
Our experiments use small, simplified tables with controlled table schemas, so the findings may not generalize to larger, noisier, or more structurally complex real world tables. Given that the throughput of the pipeline is not 100\%, some mismatched tables are dropped by different models during pipeline execution, which may impact the evaluation results. Because each model generates both the inferences and their verification code, the reported accuracy reflects the full pipeline and may be affected by verifier errors, as shown by the human evaluation. Finally, the predefined inference types and the confounding of model family, scale, and reasoning paradigm limit conclusions about unconstrained reasoning and the independent effects of these factors. Our findings should therefore be interpreted within the scope of the current experimental setting.

\section*{Acknowledgments}
We thank Ryan Dolan for the valuable discussions and insights that enhanced this work, the Illinois Computational Linguistics Lab for their supportive feedback during the project's initial stages, and the anonymous reviewers for their helpful comments.
\bibliography{custom}
\clearpage
\appendix
\onecolumn
\section{LLM Inference Pipeline Full Prompts}
\label{sec:fullprompts}

\subsection{Synthetic Table Generation Prompt}
\label{fig:table_generation_prompt}

\begin{Verbatim}[
  fontsize=\small,
  breaklines=true,
  breakanywhere=true,
  frame=single,
  framesep=4mm,
  rulecolor=\color{gray}
]
Generate exactly one synthetic statistical table in valid CSV format.

Requirements:
1. The table must have between 5 and 10 columns.
2. The table must contain exactly 15 data rows plus 1 header row.
3. Invent realistic column names, including:
   - numeric columns (integers and/or floats)
   - categorical columns
4. Numeric columns must contain only plain numbers
   (no $, commas, or % symbols).
5. Make the data plausible for a real-world dataset.
6. Ensure values vary across rows.
7. All rows must contain the same number of columns
   as the header.
8. Do not include commas inside cell values.
9. Do not include explanations, markdown, code fences,
   titles, or any text before or after the CSV.
10. Output only the CSV table.

The dataset may describe any realistic domain such as health, education, business, transportation, or demographics.
\end{Verbatim}

\bigskip

\subsection{Natural-Language Inference Generation Prompt}
\label{fig:statement_prompt}

\begin{Verbatim}[
  fontsize=\small,
  breaklines=true,
  breakanywhere=true,
  frame=single,
  framesep=4mm,
  rulecolor=\color{gray}
]
Your task is to generate 20 natural language statements that describe patterns, relationships, and notable observations in this data.

REQUIREMENTS:
1. Factual Accuracy:
   Each statement must be verifiable against the actual data. Avoid generalizations that contradict even a single record.

2. Semantic Salience:
   Focus on relationships between variables that are semantically meaningful (e.g., "women aged 21-43" or "high-income individuals"). Avoid arbitrary correlations.

3. Clarity:
   Use natural language that is easy to understand. Avoid overly complex nested conditions.

4. Variety:
   - Universal claims:
       "All X satisfy property Y"
   - Existential claims:
       "There exists at least one X that satisfies Y"
   - Majority/frequency claims:
       "Most X have property Y"
   - Negation:
       "No X where Y ..." (check subgroup max/min first)
   - Conditional claims:
       "If X, then Y"

AVOID:
- Overly specific conditions that apply to only 1-2 individuals.
- Redundant statements expressing the same fact in different ways.
- Statements that are too complex to parse (keep conditions to 2-3 attributes).
- Probabilistic language (e.g., "likely", "probably")
  unless strongly supported by the data.
\end{Verbatim}

\bigskip

\subsection{Python Verification Prompt: LLAMA-3.1-8B}
\label{fig:code_generation_prompt_LLAMA}

\begin{Verbatim}[
  fontsize=\small,
  breaklines=true,
  breakanywhere=true,
  frame=single,
  framesep=4mm,
  rulecolor=\color{gray}
]
Do NOT repeat the instructions or the code provided.
Only output the requested Python code. Do NOT wrap the code in markdown fences.
Generate a complete standalone Python script.

Create exactly one function for every input statement, named sequentially:
stmt_1, stmt_2, stmt_3, ...

The checks list must include exactly the same functions:
checks = [(1, stmt_1), (2, stmt_2), ...]

Given the statements {lines} and the table headers {df.head(0)}, write Python code using pandas to check whether each statement is True or False and print a justification.

The CSV is already located at "{full_csv_path}". Hardcode this path
directly in the script.

Convert all numeric-looking columns to integers or floats using:
pd.to_numeric(..., errors="coerce").

Each statement function should explicitly convert any numeric columns it uses before checking conditions.

Truth must always be a single Python boolean, never a pandas Series. Use .any() or .all() before bool(...) whenever an expression could return a Series.

For implication statements of the form "if A then B", always use:

antecedent = ...
consequent = ...
viol = local[antecedent & ~(consequent)]
truth = bool(viol.empty)

For boolean masks involving multiple comparisons, wrap each comparison
in parentheses:

mask = (local["x"] >= 13) & (local["y"] < 54)

Never write:

mask = local["x"] >= 13 & local["y"] < 54

Use .isin(...) directly on a Series, not on .str.

When selecting dataframe columns, always use a list, never a set.

Do not put dataframe indexing expressions directly inside f-strings.
Assign them first:

preview = viol[needed].to_dict(orient="records")
expl = f"Found {len(viol)} violating rows. Violating rows: {preview}."

Do not use numeric comparisons on categorical yes/no columns unless
the column has first been safely normalized.

Distinguish column names from category values. If a medical condition,
diagnosis, treatment type, category, group, region, gender, weather type,
or similar label appears in a statement but is not an exact column name,
do not invent a new column for it. Instead, identify the relevant
categorical column and compare against normalized string values.

For yes/no columns such as smoker, diabetes, asthma, hypertension,
or club_member, normalize the column first:

col_norm = local["smoker"].astype(str).str.strip().str.lower()
smoker_yes = col_norm.isin(["yes", "y", "true", "t", "1"])

Everything output must be valid, immediately runnable Python with no
markdown or commentary outside comments.
\end{Verbatim}

\bigskip

\subsection{Initial Python Verification Prompt: GPT-OSS}
\label{fig:initial_code_generation_prompt}

\begin{Verbatim}[
  fontsize=\small,
  breaklines=true,
  breakanywhere=true,
  frame=single,
  framesep=4mm,
  rulecolor=\color{gray}
]
System:
You are an expert data analyst.

User:
Do NOT repeat the instructions or the code provided.
Only output the requested Python code. Do NOT wrap the code
in markdown fences.

Given the following statements:

{statements_text}

and the following CSV preview showing the header row:

{table_head}

write Python code using pandas that checks whether each
statement is True or False and prints a justification.

The CSV is already located at:
"{csv_path}"

Hardcode this path directly in the script.

Requirements:
- Use pandas.
- Read the CSV from the hardcoded path.
- Convert numeric columns stored as strings into numeric values.
- Convert turn-number columns to integers when appropriate.
- Check every statement.
- Print whether each statement is True or False.
- Print a short justification for each result.
- Output only valid, immediately runnable Python.
- Do not include markdown or commentary outside Python comments.

An example Python script illustrating the expected output
format was appended to the prompt.
\end{Verbatim}

\section{Synthetic Tables}
\label{sec:synthetic_tables}
\begin{table}[htbp]
\centering
\small
\setlength{\tabcolsep}{10pt}
\renewcommand{\arraystretch}{1.15}

\begin{tabular}{|l|l|c|c|c|c|c|c|}
\hline
\textbf{patient\_id} &
\textbf{diagnosis} &
\textbf{age} &
\textbf{bp\_systolic} &
\textbf{bp\_diastolic} &
\textbf{cholesterol\_mg\_dl} &
\textbf{bmi} &
\textbf{smoker} \\
\hline
PT001001 & asthma       & 46 & 129 & 72 & 192 & 33.3 & no  \\
\hline
PT001002 & diabetes     & 62 & 143 & 82 & 223 & 22.5 & yes \\
\hline
PT001003 & diabetes     & 68 & 147 & 87 & 198 & 31.0 & no  \\
\hline
PT001004 & arthritis    & 46 & 135 & 77 & 182 & 27.6 & yes \\
\hline
PT001005 & hypertension & 76 & 119 & 74 & 245 & 22.7 & no  \\
\hline
PT001006 & arthritis    & 25 & 136 & 82 & 241 & 34.4 & no  \\
\hline
PT001007 & arthritis    & 76 & 112 & 91 & 179 & 30.0 & no  \\
\hline
PT001008 & asthma       & 28 & 130 & 83 & 185 & 26.9 & no  \\
\hline
PT001009 & arthritis    & 69 & 123 & 86 & 178 & 32.7 & no  \\
\hline
PT001010 & arthritis    & 59 & 124 & 74 & 212 & 31.2 & yes \\
\hline
PT001011 & arthritis    & 41 & 143 & 70 & 179 & 33.5 & no  \\
\hline
PT001012 & diabetes     & 24 & 127 & 98 & 237 & 33.8 & yes \\
\hline
PT001013 & migraine     & 73 & 116 & 94 & 233 & 31.2 & yes \\
\hline
PT001014 & migraine     & 56 & 122 & 78 & 232 & 32.7 & no  \\
\hline
PT001015 & diabetes     & 55 & 156 & 76 & 204 & 26.1 & no  \\
\hline
\end{tabular}%

\caption{Example of a synthetic statistical table generated by GPT-5.5}
\label{tab:table_synthetic_example}
\end{table}
\section{Example Generated Inferences and Python Checker Output}
\label{sec:inference&CheckerOutput}

\subsection{Generated Natural-Language Inferences}
\label{fig:generated_inference_example}

\noindent\textit{The following quantified inferences were generated by GPT-OSS-120B
from the synthetic patient table shown in Table~\ref{tab:table_synthetic_example}.}

\smallskip

\begin{Verbatim}[
  fontsize=\small,
  breaklines=true,
  breakanywhere=true,
  frame=single,
  framesep=4mm,
  rulecolor=\color{gray}
]
1.  All asthma patients are non-smokers.
2.  Most arthritis patients are non-smokers.
3.  There exists at least one diabetes patient who is a smoker.
4.  No hypertension patient is a smoker.
5.  If a patient is a smoker, their cholesterol level is at least 182 mg/dL.
6.  All patients with BMI greater than 33 have cholesterol of at least 179 mg/dL.
7.  No patient with cholesterol below 180 mg/dL is a smoker.
8.  All patients aged 70 or older have cholesterol above 230 mg/dL.
9.  All patients with systolic blood pressure above 140 have either diabetes or arthritis.
10. There exists at least one patient with BMI between 20 and 25 who is a smoker.
11. No patient with BMI below 25 has diastolic blood pressure above 90.
12. All patients with cholesterol above 240 mg/dL are non-smokers.
13. Most patients with BMI over 30 are non-smokers.
14. If a patient is diagnosed with migraine, their systolic blood pressure is below 130.
15. All patients with diastolic blood pressure of 70 are diagnosed with arthritis.
16. No smoker has a BMI below 22.5.
17. Most patients older than 60 have cholesterol above 200 mg/dL.
18. All patients with diastolic blood pressure of 91 are non-smokers.
19. If a patient is non-smoker, their diastolic blood pressure is at most 91.
20. All patients with cholesterol exactly 179 mg/dL are non-smokers.
\end{Verbatim}

\bigskip

\subsection{Python Checker Output}
\label{fig:checker_output_example}

\noindent\textit{The following output was generated by GPT-OSS-120B when verifying
the quantified inferences shown above against the synthetic patient table in
Table~\ref{tab:table_synthetic_example}.}

\smallskip

\begin{Verbatim}[
  fontsize=\small,
  breaklines=true,
  breakanywhere=true,
  frame=single,
  framesep=4mm,
  rulecolor=\color{gray}
]
Statement 1: TRUE
  - 1. All asthma patients are non-smokers.
  - Explanation: All 2 asthma patients are non-smokers.

Statement 2: TRUE
  - 2. Most arthritis patients are non-smokers.
  - Explanation: 66.7% of 6 arthritis patients are non-smokers.

Statement 3: TRUE
  - 3. There exists at least one diabetes patient who is a smoker.
  - Explanation: Found 2 diabetes patient(s) who smoke.

Statement 4: TRUE
  - 4. No hypertension patient is a smoker.
  - Explanation: All hypertension patients are non-smokers.

Statement 5: TRUE
  - 5. If a patient is a smoker, their cholesterol level is at least 182 mg/dL.
  - Explanation: All 5 smokers have cholesterol >=182.

Statement 6: TRUE
  - 6. All patients with BMI greater than 33 have cholesterol of at least 179 mg/dL.
  - Explanation: All 4 high-BMI patients meet the cholesterol threshold.

Statement 7: TRUE
  - 7. No patient with cholesterol below 180 mg/dL is a smoker.
  - Explanation: All low-cholesterol patients are non-smokers.

Statement 8: FALSE
  - 8. All patients aged 70 or older have cholesterol above 230 mg/dL.
  - Explanation: 1 senior patient(s) have cholesterol <=230 (ids: PT001007).
\end{Verbatim}
\begin{Verbatim}[
  fontsize=\small,
  breaklines=true,
  breakanywhere=true,
  frame=single,
  framesep=4mm,
  rulecolor=\color{gray}
]

Statement 9: TRUE
  - 9. All patients with systolic blood pressure above 140 have either diabetes or arthritis.
  - Explanation: All 4 high-systolic patients have diabetes or arthritis.

Statement 10: TRUE
  - 10. There exists at least one patient with BMI between 20 and 25 who is a smoker.
  - Explanation: Found 1 smoker(s) with BMI 20-25.

Statement 11: TRUE
  - 11. No patient with BMI below 25 has diastolic blood pressure above 90.
  - Explanation: All 2 low-BMI patients have diastolic <=90.

Statement 12: TRUE
  - 12. All patients with cholesterol above 240 mg/dL are non-smokers.
  - Explanation: All 2 high-cholesterol patients are non-smokers.

Statement 13: TRUE
  - 13. Most patients with BMI over 30 are non-smokers.
  - Explanation: 66.7% of 9 BMI>30 patients are non-smokers.

Statement 14: TRUE
  - 14. If a patient is diagnosed with migraine, their systolic blood pressure is below 130.
  - Explanation: All 2 migraine patients have systolic <130.

Statement 15: TRUE
  - 15. All patients with diastolic blood pressure of 70 are diagnosed with arthritis.
  - Explanation: All 1 patients with DBP=70 have arthritis.

Statement 16: TRUE
  - 16. No smoker has a BMI below 22.5.
  - Explanation: All 5 smokers have BMI >=22.5.

Statement 17: FALSE
  - 17. Most patients older than 60 have cholesterol above 200 mg/dL.
  - Explanation: 50.0% of 6 patients >60 have cholesterol >200.

Statement 18: TRUE
  - 18. All patients with diastolic blood pressure of 91 are non-smokers.
  - Explanation: All 1 patients with DBP=91 are non-smokers.

Statement 19: TRUE
  - 19. If a patient is non-smoker, their diastolic blood pressure is at most 91.
  - Explanation: All 10 non-smokers have diastolic <=91.

Statement 20: TRUE
  - 20. All patients with cholesterol exactly 179 mg/dL are non-smokers.
  - Explanation: All 2 patients with cholesterol 179 are non-smokers.
\end{Verbatim}

\section{Metrics}
\label{sec:metrics}
\begin{table}[htbp]
\centering
\includegraphics[width=1\linewidth]{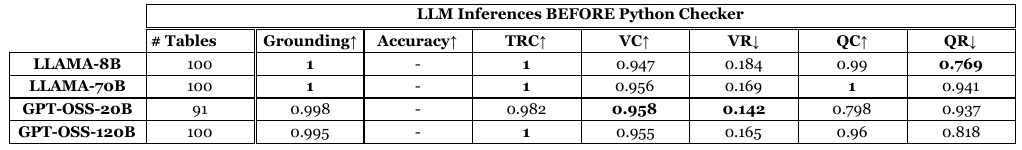}
\caption{Performance of open-weight LLMs on \textsc{Stat-to-Text} generated inferences' grounding, coverage (TRC, VC, QC), and diversity (VR, QR) \textit{before} Python-based verification. Accuracy is non-existent before Python inferences checker. Metrics annotated with $\uparrow$ indicate that higher values correspond to better performance, whereas metrics annotated with $\downarrow$ indicate that lower values are preferred.}
\label{tab:step-b-results-overall}
\end{table}

\newpage
\section{Examples of Disagreements Between the Python Checker and Human Annotation}
\label{sec:human_eval}


\noindent\textbf{\large Example 1: Correct Row Selection with an Incorrect Verdict}
\label{sec:human_eval_1}

\smallskip

\setlength{\fboxsep}{8pt}
\noindent\fcolorbox{gray}{gray!5}{%
\begin{minipage}{0.95\linewidth}

\textbf{Model:} LLAMA-3.1-8B

\medskip

\textbf{Statement.}
There exists at least one farm that grows corn and has a fertilizer
application of less than 645 kg.

\medskip

\begin{tabular}{@{}ll@{}}
\textbf{Checker result:} & \texttt{FALSE} \\
\textbf{Human label:}    & \texttt{TRUE}
\end{tabular}

\medskip

\textbf{Checker explanation.}
{\small\ttfamily
Found 1 violating rows. Violating rows:
[\{'crop\_type': 'corn', 'fertilizer\_kg': 611\}].
}

\medskip

\textbf{Error analysis.}
The checker identifies a row containing a corn farm with a fertilizer
application of 611 kg, which satisfies the existential claim. However, it
incorrectly treats this satisfying row as a violation and assigns a false
verdict.

\end{minipage}%
}

\bigskip
\bigskip


\noindent\textbf{\large Example 2: Majority Quantifier Evaluated as Universal}
\label{sec:human_eval_2}

\smallskip

\noindent\fcolorbox{gray}{gray!5}{%
\begin{minipage}{0.95\linewidth}

\textbf{Model:} LLAMA-3.1-8B

\medskip

\textbf{Statement.}
Most employees have a performance rating of 4.0 or higher.

\medskip

\begin{tabular}{@{}ll@{}}
\textbf{Checker result:} & \texttt{FALSE} \\
\textbf{Human label:}    & \texttt{TRUE}
\end{tabular}

\medskip

\textbf{Checker explanation.}
{\small\ttfamily
Found 1 violating rows. Violating rows:
[\{'employee\_id': 'E019002', 'performance\_rating': 3.8\}].
}

\medskip

\textbf{Error analysis.}
Only one employee has a performance rating below 4.0, so the majority
of employees satisfy the stated threshold. The checker incorrectly
rejects the statement because it finds one non-satisfying employee,
effectively evaluating \textit{most} as if it were the universal
quantifier \textit{all}.

\end{minipage}%
}

\bigskip
\bigskip


\noindent\textbf{\large Example 3: Verdict Contradicting the Checker Explanation}
\label{sec:human_eval_3}

\smallskip

\noindent\fcolorbox{gray}{gray!5}{%
\begin{minipage}{0.95\linewidth}

\textbf{Model:} LLAMA-3.1-8B

\medskip

\textbf{Statement.}
For all individuals with a diagnosis of hypertension, their age is
greater than 40.

\medskip

\begin{tabular}{@{}ll@{}}
\textbf{Checker result:} & \texttt{FALSE} \\
\textbf{Human label:}    & \texttt{TRUE}
\end{tabular}

\medskip

\textbf{Checker explanation.}
{\small\ttfamily
Found 0 violating rows. Violating rows: [].
}

\medskip

\textbf{Error analysis.}
The checker reports that no violating rows were found, which supports the
universal statement. Nevertheless, it outputs a false verdict, directly
contradicting its own explanation.

\end{minipage}%
}

\bigskip
\bigskip

\newpage
\noindent\textbf{\large Example 4: Disjunction Evaluated Incorrectly}
\label{sec:human_eval_4}

\smallskip

\noindent\fcolorbox{gray}{gray!5}{%
\begin{minipage}{0.95\linewidth}

\textbf{Model:} LLAMA-3.1-8B

\medskip

\textbf{Statement.}
All players in the table are either guards, centers, or forwards.

\medskip

\begin{tabular}{@{}ll@{}}
\textbf{Checker result:} & \texttt{FALSE} \\
\textbf{Human label:}    & \texttt{TRUE}
\end{tabular}

\medskip

\textbf{Checker explanation.}

{\small\ttfamily
Found 15 violating rows. Violating rows:
[\{'position': 'guard'\}, \{'position': 'guard'\},
\{'position': 'guard'\}, \{'position': 'center'\},
\{'position': 'center'\}, \{'position': 'forward'\},
\{'position': 'forward'\}, \{'position': 'guard'\},
\{'position': 'forward'\}, \{'position': 'forward'\},
\{'position': 'guard'\}, \{'position': 'guard'\},
\{'position': 'guard'\}, \{'position': 'forward'\},
\{'position': 'center'\}].
}

\medskip

\textbf{Error analysis.}
Every row contains one of the three permitted position labels---\texttt{guard},
\texttt{center}, or \texttt{forward}---so no row violates the disjunctive
condition. Nevertheless, the checker marks all 15 rows as violations and
assigns a false verdict.

\end{minipage}%
}

\newpage
\FloatBarrier
\end{document}